\documentclass[lettersize,journal]{IEEEtran}
\usepackage{amsmath,amssymb,amsfonts}
\usepackage{algorithmic}
\usepackage{algorithm}
\usepackage{array}
\usepackage[caption=false,font=normalsize,labelfont=sf,textfont=sf]{subfig}
\usepackage{textcomp}
\usepackage{stfloats}
\usepackage{url}
\usepackage{verbatim}
\usepackage{graphicx}
\usepackage{cite}
\usepackage{xcolor}
\usepackage{bm}
\usepackage{booktabs}
\usepackage{multirow}
\usepackage{soul}
\usepackage{enumitem}
\usepackage[utf8]{inputenc}
\usepackage{ntheorem}
\usepackage{wrapfig}

\DeclareUnicodeCharacter{2113}{\ell}

\usepackage{xcolor} 
\usepackage{hyperref}
\hypersetup{hypertex=true,
	colorlinks=true,
	linkcolor=blue,
	anchorcolor=blue,
	citecolor=blue}
\def\BibTeX{{\rm B\kern-.05em{\sc i\kern-.025em b}\kern-.08em
		T\kern-.1667em\lower.7ex\hbox{E}\kern-.125emX}}
\begin{document}

\title{Single Image Reflection Removal via Dual-Prior Interaction Transformer
}

	
\author{Yue~Huang, Tianle~Hu,  Yu~Chen, Ziang~Li, Jie~Wen,~\IEEEmembership{Senior Member,~IEEE}, and Xiaozhao~Fang*
	
	\IEEEcompsocitemizethanks{
		\IEEEcompsocthanksitem 
		Yue Huang, Yu Chen, Ziang Li, and Xiaozhao Fang are with the School of Automation, Guangdong University of Technology, Guangzhou 510006, China (e-mail: 17324004911@163.com,  chenyu9265324@163.com, 2112404083@mail2.gdut.edu.cn, xzhfang168@126.com).
		\IEEEcompsocthanksitem Tianle~Hu is with the School of Computer Science and Technology, Guangdong University of Technology, Guangzhou 510006, China (e-mail: hutianlegdut@163.com)
		\IEEEcompsocthanksitem 
		Jie Wen is with the Shenzhen Key Laboratory of Visual Object Detection and Recognition, Harbin Institute of Technology, Shenzhen, Shenzhen 518055, China (e-mail: jiewen\_pr@126.com).
		\IEEEcompsocthanksitem 
		Corresponding author: Xiaozhao Fang (xzhfang168@126.com).
	}

}

\markboth{IEEE Transactions on Image Processing}%
{Shell \MakeLowercase{\textit{et al.}}: A Sample Article Using IEEEtran.cls for IEEE Journals}


\maketitle

\begin{abstract}
	To address the challenge of recovering transmission content from the limited information in a single mixed image, recent methods introduce various priors as supplementary guidance, including general priors from pre-trained models and task-oriented priors such as text prompts and reflection estimation. However, these priors provide only coarse-grained perception of transmission content, limiting restoration effectiveness. To address this issue, we propose a Dual-Prior Interaction Transformer (DPIT) that introduces a fine-grained transmission prior and fuses it with general prior for enhanced restoration guidance. Specifically, we design a Local Linear Correction Network (LLCN) to generate transmission prior through pixel-wise scaling and bias operations. Compared to direct pixel generation methods, this approach achieves significant performance gains under the same low-parameter budget. To enable effective dual-prior interaction, we propose a Dual-Stream Channel Reorganization Transformer (DSCRT) based on a layer separation architecture. The DSCRT employs the Dual-Stream Channel Reorganization Attention Mechanism (DSCRAM), which leverages the complementarity of heterogeneous features and the exclusivity of layer separation objectives to construct a simplified yet effective computational target. DSCRAM reorganizes the dual-stream structure at the channel level to ensure both streams contain heterogeneous features before attention computation. By leveraging the complementarity of heterogeneous features and the exclusivity of layer separation objectives, the attention mechanism effectively performs intra-stream feature separation and cross-stream feature complementation. Extensive experiments validate that DPIT achieves state-of-the-art performance across multiple benchmark datasets.
\end{abstract}



\section{Introduction}\label{sec:introduction}

\IEEEPARstart{W}{hen} image acquisition occurs through transparent media such as glass, the captured images typically exhibit reflection degradation. This phenomenon adversely affects the performance of numerous vision technologies, including object detection, scene understanding, and depth estimation, consequently limiting their effectiveness in practical applications such as mobile photography, video surveillance, autonomous driving, and industrial inspection \cite{simon2015reflection}, \cite{li2021deep}, \cite{guan2023defect}. To mitigate this problem, diverse reflection removal approaches have been developed, which can be categorized into multi-image methods \cite{sarel2004separating}, \cite{agrawal2005removing},  \cite{guo2014robust}, \cite{han2017reflection}, \cite{sinha2012image}, \cite{li2013exploiting}, \cite{xue2015computational}, \cite{yang2016robust}, polarization-based techniques \cite{lei2020polarized}, intrinsic image solutions \cite{levin2007user}, auxiliary data-driven approaches \cite{zhong2024language}, and single image methods \cite{fan2017generic}, \cite{zhang2018single}, \cite{wei2019single}. Among these, single image reflection removal demonstrates notable practical advantages, as it operates on a single input without requiring specialized hardware, user interaction, or auxiliary information. Given these merits, single image reflection removal has emerged as a significant and actively pursued research field in computer vision.

Nevertheless, due to the limited information available in a single mixed image, existing methods often struggle to accurately recover the transmission content, frequently yielding outputs with residual artifacts, color distortions, or incomplete reflection removal. To address this information insufficiency, various prior-based approaches have been developed, evolving from hand-crafted priors such as gradient sparsity and relative smoothness \cite{li2014single} to learned priors derived from pre-trained models \cite{fan2017generic}, \cite{yang2018seeing}, \cite{zhang2018single}, \cite{wan2018crrn}. More recently, task-specific priors including textual guidance \cite{zhong2024language} and explicit reflection estimation \cite{li2023two} have emerged. However, these priors only provide coarse perception of the transmission content within the mixed image, limiting the effectiveness of the information they supplement for reflection removal.

Addressing the limitation that existing prior designs only provide coarse-grained perception of transmission content, we propose leveraging reflection removal results from a pre-trained dedicated network as transmission prior to provide fine-grained guidance for another network's reflection removal process. This direction builds upon recent advances that recognize the value of explicit transmission modeling. YTMT \cite{hu2021trash} employs a two-stage architecture where initial transmission estimates guide subsequent refinement, while RDNet \cite{zhao2025reversible} incorporates transmission-rate-aware prompts to modulate feature extraction. These pioneering efforts validate the potential of utilizing transmission prior to enhance reflection removal performance.

However, while directly employing existing reflection removal networks for prior generation appears straightforward, it introduces a fundamental trade-off. High-performance networks typically require substantial parameters and computational resources, limiting the design flexibility of subsequent networks, whereas lightweight alternatives fail to provide sufficiently accurate guidance. To resolve this challenge, we reconceptualize the physical modeling of reflection superimposition to achieve efficient yet effective transmission prior generation. Existing physical formulations have evolved from simple linear model $I = aT + R$ \cite{wan2018crrn}, \cite{li2020single}, \cite{yang2018seeing}, where $I$ denotes the observed image, $T$ represents the transmission layer, $R$ is the reflection layer, and $a$ is a scalar attenuation coefficient, to sophisticated non-linear model $I = W \odot T + (1 - W) \odot R$ \cite{wen2019single}, \cite{dong2021location}, and further to component synergy frameworks $I = T + R + \Phi(T, R)$ \cite{hu2023single} with $\Phi(T, R)$ modeling the residual term. While varying in complexity, these models uniformly formulate the problem as direct pixel generation, necessitating substantial parameters to estimate pixel values from scratch. We propose a paradigmatic shift from pixel generation to pixel selection through the Local Linear Correction Model (LLCM), formulated as $T = sI + b$, where $s$ and $b$ denote learnable per-pixel scaling and bias terms respectively. By requiring only transformation parameters rather than direct pixel estimation, this approach achieves superior reflection removal performance under limited parameter budgets while maintaining reconstruction quality.

Since transmission prior are trained on limited public reflection removal datasets, their generalization capability is inherently limited. To address this constraint, we incorporate general prior from pre-trained model as complementary guidance. For effective fusion of transmission and general prior, we draw inspiration from dual-stream feature interaction mechanisms in layer separation architectures that address similar feature fusion challenges. These mechanisms have evolved from activation-based inter-stream selection to convolutional gating for channel dependencies and recently to self-attention for global correlations. While progressively improving interaction quality, this evolution escalates both computational and structural complexity. For instance, DSIT \cite{hu2024single} concatenates dual-stream features along batch and sequence dimensions to enable simultaneous dual-stream self-attention and cross-attention, resulting in considerable overhead.

To mitigate the computational burden of attention-based dual-stream methods, we recognize that the complementarity of dual-prior features and the exclusivity of layer separation objectives enable simpler attention target construction. Through dual-stream channel reorganization, we construct targets wherein both streams contain heterogeneous features at the channel level. This design allows the attention mechanism to efficiently execute intra-stream separation and cross-stream complementation within the generation stream, while the exchange stream provides guided complementary information. Based on this framework, we propose the DSCRAM and the DSCRAT for dual-prior integration, feature interaction, and layer separation, achieving superior transmission quality with substantially reduced computational complexity.

In summary, our work makes the following contributions:

\begin{enumerate}
	\item We propose DPIT, a novel dual-prior interaction approach for reflection removal that leverages transmission prior. By introducing LLCN for lightweight transmission prior generation and designing DSCRAT for effective dual-prior interaction and layer separation, DPIT achieves state-of-the-art performance across multiple benchmarks.
	
	\item We propose LLCN for lightweight transmission prior generation, based on LLCM formulated as $T = sI + b$ where $s$ and $b$ are learnable scaling and bias parameters. Through a methodological shift from pixel generation to pixel selection, LLCM enables LLCN to achieve superior transmission prior generation performance under limited parameter budgets.
	
	\item We propose DSCRAT for dual-prior interaction and layer separation. By leveraging the complementarity of heterogeneous features and the exclusivity of layer separation objectives, DSCRAT reorganizes the dual-stream structure such that both streams benefit from heterogeneous features. This reconstructed target structure enables the attention mechanism to effectively perform intra-stream feature separation and cross-stream feature complementation, achieving superior reflection removal performance with significantly reduced computational costs.
\end{enumerate}

\section{Related Work}\label{sec:related_work}

\subsection{Prior Construction}

Prior information plays a crucial role in reflection removal by supplementing the limited information available from a single mixed image. Current methods obtain prior information through various approaches, which can be broadly categorized into two types: general priors and task-specific priors.

General priors leverage pre-trained models to extract semantically rich features. Zhang et al.~\cite{zhang2018single} pioneer the extraction of hypercolumn features from pre-trained VGG-19~\cite{simonyan2014very}, enforcing layer separation through perceptual loss. Hu et al.~\cite{hu2024single} employ pre-trained Swin Transformer~\cite{liu2021swin} to capture global features and enhance inter-layer correlation modeling via dual attention mechanisms. Zhao et al.~\cite{zhao2025reversible} utilize pre-trained FocalNet features to construct a reversible decoupling network. While general priors provide rich semantic representations, they are derived from generic visual recognition tasks and lack specific modeling of the reflection formation process, limiting their effectiveness in guiding accurate layer separation.

To overcome this limitation, task-specific priors tailored for reflection removal have been developed. For geometric priors, Wan et al.~\cite{wan2018crrn} employ multi-scale guidance for concurrent reflection removal, while He et al.~\cite{he2025rethinking} leverage depth information to establish spatial geometric constraints for reflection separation. For physical priors, Lei et al.~\cite{lei2020polarized} exploit polarization characteristics to capture differences in light polarization states across transmission and reflection components, and Wan et al.~\cite{wan2020reflection} introduce scene depth as a physical cue to distinguish reflection from transmission. For gradient priors, Fan et al.~\cite{fan2017generic} utilize edge sparsity constraints to promote gradient exclusivity between layers. For semantic priors, Zhong et al.~\cite{zhong2024language} utilize language descriptions to provide scene-level guidance for the separation process. For reflection priors, Dong et al.~\cite{dong2021location} propose a location-aware approach that estimates reflection confidence maps to guide removal, Li et al.~\cite{li2023two} introduce RAGNet which estimates the reflection layer first and uses its features to guide transmission recovery, and Zhu et al.~\cite{zhu2024revisiting} develop a maximum reflection filter to identify and characterize reflection locations for targeted removal.

Among task-specific priors, transmission prior have been gradually proven effective for layer separation by providing fine-grained perception and explicit constraints on the final reflection removal results. Hu and Guo~\cite{hu2021trash} introduce YTMT with a transmission refinement network in the second stage, achieving progressive optimization through selective feature propagation. Zhao et al.~\cite{zhao2025reversible} propose a transmittance-aware prompt generator that dynamically modulates network features through transmission rate parameters, adapting to varying reflection intensities and substantially improving model robustness across diverse scenarios.

To further enhance transmission prior, integrating general prior from pre-trained model as complementary information becomes essential, which necessitates effective heterogeneous feature interaction mechanisms. Hu and Guo~\cite{hu2021trash} address this in YTMT by assessing information utility through activation functions and exchanging low-value information between dual streams. Building upon this foundation, Hu and Guo~\cite{hu2023single} introduce DSRNet with a mutually-gated interaction (MuGI) mechanism that enables more efficient bidirectional information exchange to capture component synergy. Subsequently, Hu et al.~\cite{hu2024single} develop DSIT, which devises a dual-attention interactive structure that includes dual-stream self-attention and layer-aware dual-stream cross-attention to simultaneously capture intra-layer and inter-layer feature correlations. Furthermore, DSIT modulates single-stream pre-trained Transformer embeddings with dual-stream convolutional features through cross-architecture interactions to provide richer semantic priors, thereby advancing separation performance.

\subsection{Physical Model Construction}

Physical modeling establishes explicit optimization objectives for reflection removal by characterizing the reflection superposition process. Early work adopted the linear additive model $I=T+R$. Levin and Weiss~\cite{levin2007user} combine this model with gradient sparsity priors for layer separation, while Fan et al.~\cite{fan2017generic} incorporate it into deep learning frameworks. However, this model neglects spatial variations in reflection intensity.

To address this, Wan et al.~\cite{wan2018crrn} introduce weighted coefficients, proposing $I=\alpha T+\beta R$ with independent attenuation for each layer to account for real-world degradation effects. Zhang et al.~\cite{zhang2018single} enhance separation by introducing exclusion loss to enforce gradient independence between layers.

Beyond scalar coefficients, pixel-wise alpha matting approaches offer finer control. Wen et al.~\cite{wen2019single} propose $I=W\odot T+(1-W)\odot R$ where $W$ represents a spatially-varying alpha matte, enabling localized modeling of reflection intensity. Dong et al.~\cite{dong2021location} adopt a similar formulation to emphasize location-aware reflection characteristics.

Subsequent work has incorporated physical degradation effects to more accurately characterize reflection phenomena. Wan et al.~\cite{wan2017benchmarking} introduce blur kernel modeling through $I=T+k\otimes R$ to capture defocus effects. For ghosting phenomena in thick glass, Shih et al.~\cite{shih2015reflection} introduce a double-kernel model $I=T+k_1\otimes R+k_2\otimes R$ to capture spatial displacement from multiple reflections at different glass surfaces. Zheng et al.~\cite{zheng2021single} account for energy attenuation by proposing an absorption-aware model $I=e\cdot T+\Phi\cdot R$, where $e$ and $\Phi$ relate to glass material properties and absorption characteristics.

With deeper understanding of complex real-world scenes, approaches beyond linear assumptions have emerged. Wen et al.~\cite{wen2019single} propose a fully nonlinear model learned through neural networks, capturing complex interactions. Balancing flexibility and interpretability, Hu and Guo~\cite{hu2023single} introduce a residual enhancement model $I=T+R+\Phi(T,R)$, decomposing the mixing process into linear components and nonlinear residuals.

\begin{figure*}[t]
	\centering
	\includegraphics[width=1\linewidth]{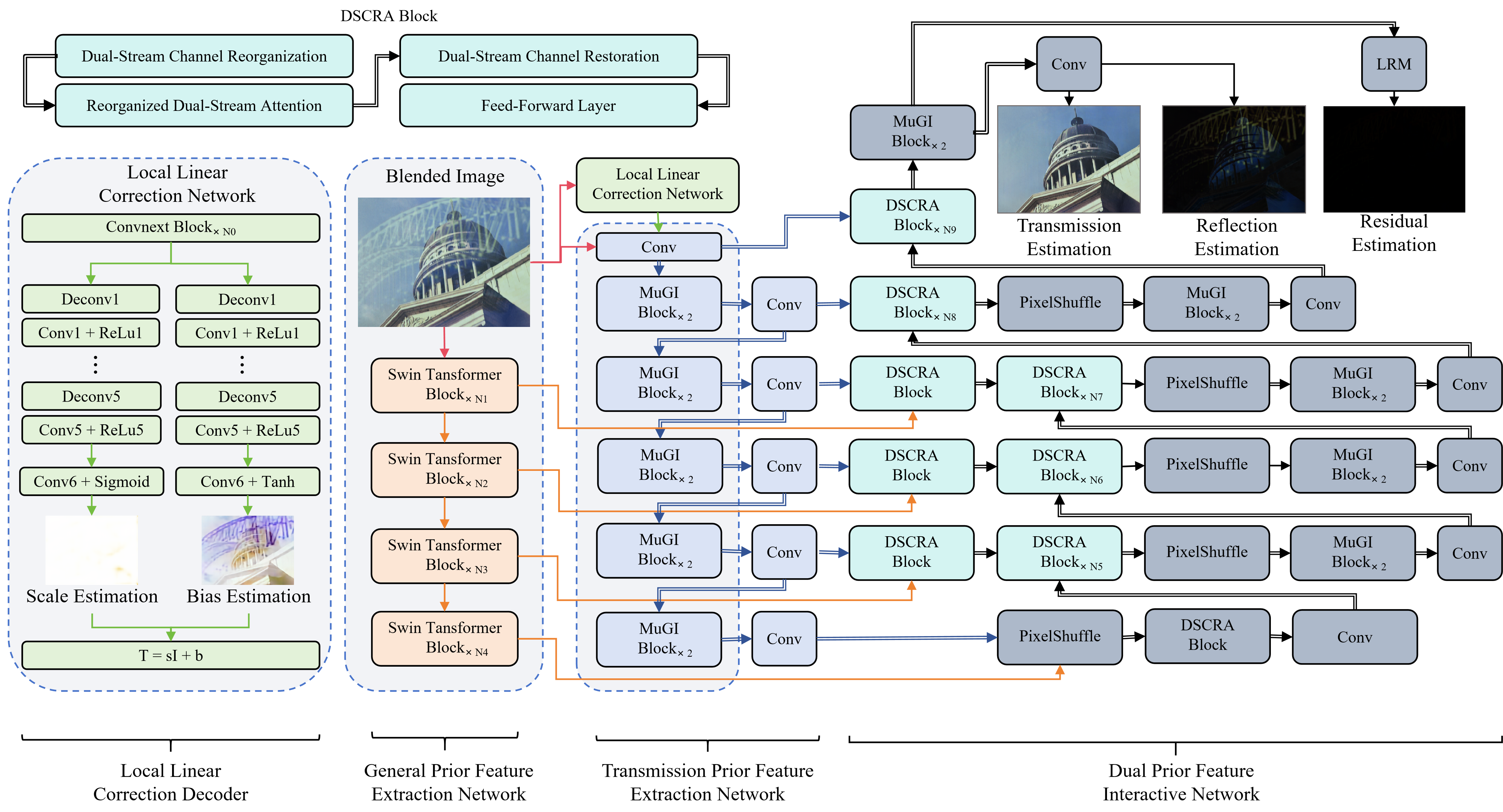}
\caption{Architecture of the Dual-Prior Interaction Transformer (DPIT), including General Prior Feature Extraction Network, Transmission Prior Feature Extraction Network (TPFEN), Local Linear Correction Network (LLCN), Dual-Prior Feature Interaction Network (DPFIN), and Local Linear Correction Network (LLCN). Double-line segments represent dual-stream paths. LRM denotes the Learnable Residue Module \cite{hu2024single}. DSCRA Block (DSCRAB) represents the proposed Dual-Stream Channel Reorganization Attention Block. MuGI Block refers to the Mutually-Gated Interactive Block \cite{hu2023single}.}	
	\label{fig:example2}	
\end{figure*}

\section{Proposed Method}\label{sec:method}

\subsection{Overall Architecture}

As illustrated in Fig.~\ref{fig:example2}, the proposed DPIT consists of four core components, namely General Prior Feature Extraction Network (GPFEN), Local Linear Correction Network (LLCN), Transmission Prior Feature Extraction Network (TPFEN), and Dual-Prior Feature Interaction Network (DPFIN). 

Given a mixed image $I$, GPFEN extracts multi-scale general prior features, while LLCN generates the transmission prior $\hat{T}_{\text{prior}}$. TPFEN then processes both $I$ and $\hat{T}_{\text{prior}}$ to obtain multi-scale transmission prior features. DPFIN subsequently employs Dual-Stream Channel Reorganization Attention Block (DSCRAB) for dual-prior interaction through same-layer and cross-layer fusion, outputting the transmission layer $\hat{T}$, reflection layer $\hat{R}$, and residual term $\hat{\Phi}$. It should be noted that DSCRT is a variant of DPIT without the LLCN component. The following sections will provide detailed descriptions of LLCN, DSCRAB, and DSCRAT.

\subsection{Local Linear Correction Model}

As illustrated in Fig.~\ref{fig:example2}, the proposed local linear correction model focuses on achieving efficient transmission layer prior generation. This model reformulates transmission layer estimation as an adaptive linear correction problem applied to the blended image:
\begin{equation}
	\hat{T}_{\text{prior}} = sI + b,
\end{equation}
where $I \in \mathbb{R}^{3 \times H \times W}$ denotes the blended image, $s \in \mathbb{R}^{3 \times H \times W}$ and $b \in \mathbb{R}^{3 \times H \times W}$ represent pixel-wise scaling factors and bias terms across channels, respectively, with $H$ and $W$ denoting the image height and width, respectively.

We construct the Local Linear Correction Network (LLCN) based on this model. The network employs a pre-trained ConvNeXt-Base~\cite{liu2022convnet} as its feature extraction backbone, extracting deep semantic features $F \in \mathbb{R}^{1024 \times 7 \times 7}$ from the input image. These features are then fed into two parallel decoders with identical architectures. Each decoder progressively upsamples the features to match the original input resolution through cascaded modules comprising deconvolution, convolution, and ReLU activation layers. The two decoders are used to generate the scaling factors and bias terms, respectively:

\begin{equation}
	s = \text{Sigmoid}(\text{Decoder}_1(F)), \quad b = \text{Tanh}(\text{Decoder}_2(F)),
\end{equation}

where the Sigmoid activation constrains $s$ to $[0,1]$ for pixel intensity modulation, and the Tanh activation constrains $b$ to $[-1,1]$ for brightness deviation correction.

Our approach is based on selection rather than generation. Unlike end-to-end methods that directly regress the transmission layer, LLCN learns pixel-level selection strategies from the blended image: $s$ controls the preservation or suppression of local intensities, while $b$ compensates for brightness offsets introduced by reflections. This parameterization simplifies the learning objective from complete transmission layer reconstruction to effective information selection from the blended image, significantly reducing model complexity. Furthermore, by fine-tuning the pre-trained ConvNeXt-Base, the network leverages generic visual priors learned from large-scale datasets, thereby enhancing prior generation quality.

The training objective employs mean squared error loss:
\begin{equation}
	\mathcal{L}_{\text{correction}} = \frac{1}{N}\|sI + b - T_{\text{gt}}\|_2^2,
\end{equation}
where $T_{\text{gt}}$ denotes the ground-truth transmission layer, and $N = 3HW$ represents the total number of image elements.

\begin{figure*}[t]
	\centering
	\includegraphics[width=1\linewidth]{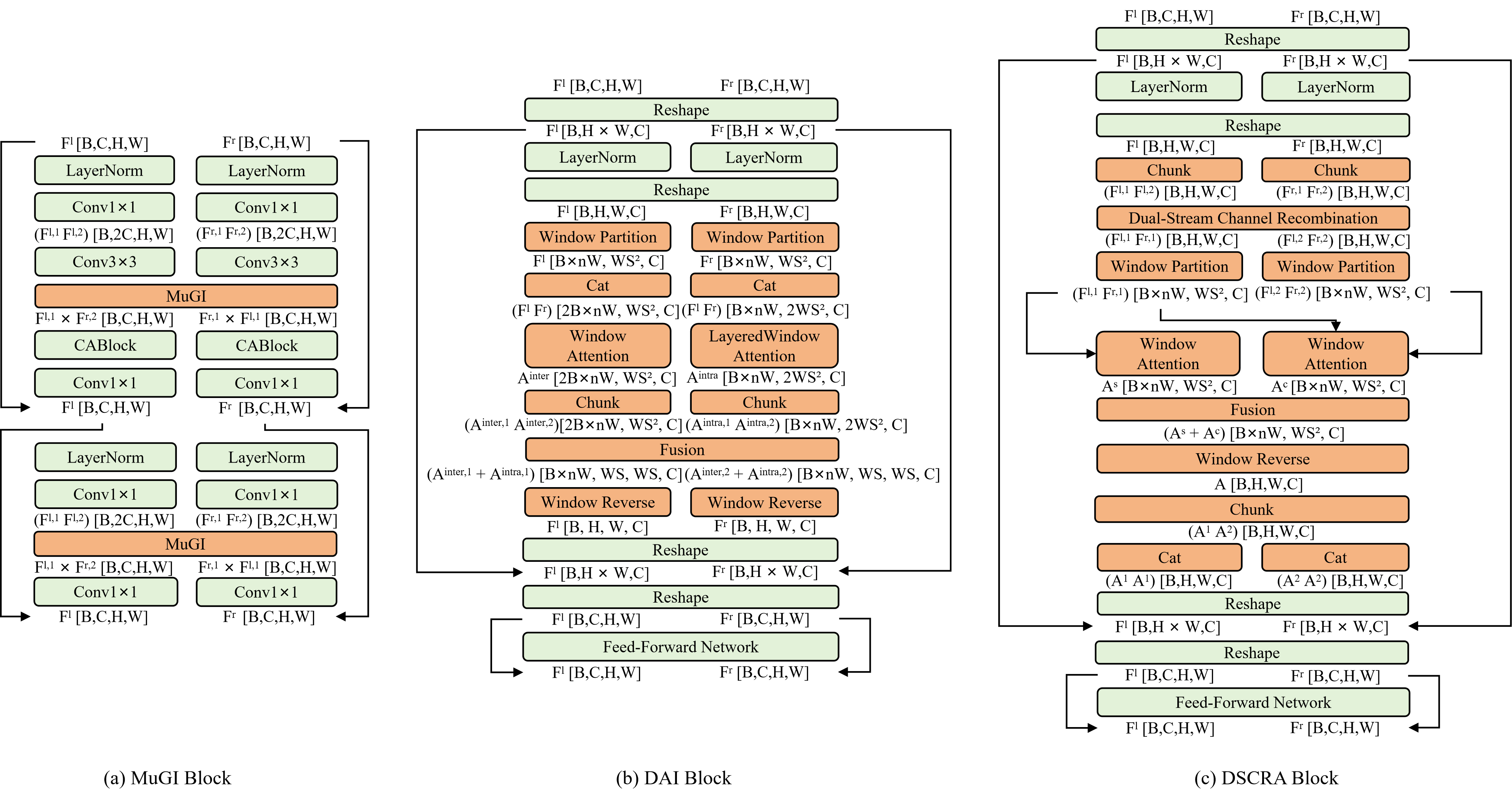}
\caption{Comparison of three dual-stream interaction modules: 
	(a) Mutually-Gated Interactive (MuGI) Block~\cite{hu2023single}, 
	(b) Dual-Attention Interaction (DAI) Block~\cite{hu2024single}, and 
	(c) Dual-Stream Channel Reorganization Attention Block (DSCRAB). 
	$F^l$ and $F^r$ denote the left and right stream features, respectively. 
	$A^{\text{intra}}$ represents intra-layer attention, 
	$A^{\text{inter}}$ represents inter-layer attention, 
	$A^s$ represents intra-stream self-attention, and 
	$A^c$ represents cross-stream cross-attention. 
	nW denotes the number of windows, and WS denotes the window size.}
	\label{fig:example1}	
\end{figure*}

\subsection{Dual-Stream Channel Reorganization Attention}

As depicted in Fig.~\ref{fig:example1}, the Dual-Stream Channel Reorganization Attention Block, referred to as DSCRAB, realizes cross-prior feature interaction and dual-stream decomposition through an elegant combination of channel reorganization and window attention mechanisms. The module takes as input the left-stream features $F^l$ and right-stream features $F^r$, where $F^l, F^r \in \mathbb{R}^{B \times C \times H \times W}$. Initially, both feature sets are reshaped to $\mathbb{R}^{B \times N \times C}$ with $N = H \times W$, and stored as $F_{\text{skip}}^l, F_{\text{skip}}^r$ to enable subsequent residual connections. Following layer normalization, they are reshaped to $\mathbb{R}^{B \times H \times W \times C}$ and denoted as $\tilde{F}^l$ and $\tilde{F}^r$, respectively. The features are subsequently partitioned into two halves along the channel dimension as follows.
\begin{equation}
	[\tilde{F}_1^l, \tilde{F}_2^l] = \text{Chunk}(\tilde{F}^l), \quad [\tilde{F}_1^r, \tilde{F}_2^r] = \text{Chunk}(\tilde{F}^r),
\end{equation}
where $\tilde{F}_1^l, \tilde{F}_2^l, \tilde{F}_1^r, \tilde{F}_2^r \in \mathbb{R}^{B \times H \times W \times \frac{C}{2}}$. Through cross-stream concatenation, we construct the generation stream and exchange stream as follows.
\begin{equation}
	F_{\text{gen}} = \text{Cat}(\tilde{F}_1^l, \tilde{F}_1^r), \quad F_{\text{exch}} = \text{Cat}(\tilde{F}_2^l, \tilde{F}_2^r).
\end{equation}

The generation stream aggregates the first-half channels from both priors, while the exchange stream retains the second-half channels, both restored to the full dimensionality of $\mathbb{R}^{B \times H \times W \times C}$. Subsequently, the Window Partition operation spatially divides both the generation and exchange streams into non-overlapping local windows $F_{\text{gen}}^{\text{win}}, F_{\text{exch}}^{\text{win}} \in \mathbb{R}^{B \times N_w \times M \times C}$, where $N_w$ denotes the total number of windows and $M$ represents the token count within each window.

DSCRAB incorporates two parallel window attention modules. Both modules derive their queries from the generation stream, which serves as the dominant information source for computing intra-stream self-attention and cross-stream attention, respectively. 

For intra-stream self-attention, queries, keys, and values are all computed exclusively from the generation stream through:
\begin{equation}
	Q_{\text{intra}} = F_{\text{gen}}^{\text{win}}W_{q1}, \quad K_{\text{intra}} = F_{\text{gen}}^{\text{win}}W_{k1}, \quad V_{\text{intra}} = F_{\text{gen}}^{\text{win}}W_{v1}.
\end{equation}
\begin{equation}
	A_{\text{intra}} = \text{SoftMax}\left(\frac{Q_{\text{intra}}K_{\text{intra}}^{\top}}{\sqrt{D}} + B_{\text{intra}}\right)V_{\text{intra}}.
\end{equation}

In contrast, the cross-stream attention module generates queries from the generation stream while extracting keys and values from the exchange stream.
\begin{equation}
	Q_{\text{cross}} = F_{\text{gen}}^{\text{win}}W_{q2}, \quad K_{\text{cross}} = F_{\text{exch}}^{\text{win}}W_{k2}, \quad V_{\text{cross}} = F_{\text{exch}}^{\text{win}}W_{v2}.
\end{equation}
\begin{equation}
	A_{\text{cross}} = \text{SoftMax}\left(\frac{Q_{\text{cross}}K_{\text{cross}}^{\top}}{\sqrt{D}} + B_{\text{cross}}\right)V_{\text{cross}}.
\end{equation}

Here, $W_{q1}, W_{k1}, W_{v1}, W_{q2}, W_{k2}, W_{v2}$ denote learnable projection matrices, $D$ represents the feature dimension, and $B_{\text{intra}}, B_{\text{cross}}$ are relative position biases. The attention outputs $A_{\text{intra}}, A_{\text{cross}} \in \mathbb{R}^{BN_w \times M \times C}$ capture different aspects of feature relationships. The intra-stream self-attention mechanism captures long-range dependencies within the generation stream, while the cross-stream attention establishes explicit correspondences between the generation and exchange streams. This dual-attention architecture, with the generation stream as its primary driver, accomplishes cross-prior channel reorganization and provides a solid foundation for effectively redistributing features back to the original dual-stream structure in subsequent stages.

The outputs from both attention pathways are aggregated and processed through the Window Reverse operation to restore the spatial structure.
\begin{equation}
	F_{\text{combined}} = \text{WindowReverse}(A_{\text{intra}} + A_{\text{cross}}),
\end{equation}
where $F_{\text{combined}} \in \mathbb{R}^{B \times H \times W \times C}$. The fused features are then bisected along the channel dimension through the Chunk operation.
\begin{equation}
	[F_{\text{out}}^l, F_{\text{out}}^r] = \text{Chunk}(F_{\text{combined}}),
\end{equation}
where $F_{\text{out}}^l, F_{\text{out}}^r \in \mathbb{R}^{B \times H \times W \times \frac{C}{2}}$. Through the Cat operation, each component is duplicated and concatenated to restore the dual-stream architecture with full channel capacity.
\begin{equation}
	F^{l,\text{attn}} = \text{Cat}(F_{\text{out}}^l, F_{\text{out}}^l), \quad F^{r,\text{attn}} = \text{Cat}(F_{\text{out}}^r, F_{\text{out}}^r),
\end{equation}
where $F^{l,\text{attn}}, F^{r,\text{attn}} \in \mathbb{R}^{B \times H \times W \times C}$. The dual-stream features are reshaped to $\mathbb{R}^{B \times N \times C}$ and undergo the first residual connection with their original inputs.

\begin{equation}
	F^{l,\text{res1}} = F_{\text{skip}}^l + \alpha F^{l,\text{attn}}, \quad F^{r,\text{res1}} = F_{\text{skip}}^r + \alpha F^{r,\text{attn}},
\end{equation}
where $\alpha$ is a learnable scaling factor. The residually connected features are reshaped back to $\mathbb{R}^{B \times C \times H \times W}$ and forwarded to a feed-forward network for further refinement. This feed-forward network comprises two critical components that work in tandem. The gating interaction module facilitates selective information exchange between the left and right streams through adaptive gating mechanisms. Meanwhile, the channel attention module performs dynamic modulation of individual channel responses. The feed-forward network output undergoes a second residual connection.

\begin{equation}
	F^{l,\text{final}} = F^{l,\text{res1}} + \beta \text{FFN}(F^{l,\text{res1}}),
\end{equation}
\begin{equation}
	F^{r,\text{final}} = F^{r,\text{res1}} + \beta \text{FFN}(F^{r,\text{res1}}),
\end{equation}
where $\beta$ is a learnable scaling factor, and $F^{l,\text{final}}, F^{r,\text{final}} \in \mathbb{R}^{B \times C \times H \times W}$ represent the final outputs. Through the synergistic interplay of channel reorganization, dual window attention modules, and the feed-forward network, DSCRAB achieves sophisticated feature interaction between the general prior and transmission prior while preserving the independence of dual streams, thereby providing complementary constraints that facilitate the subsequent separation of transmission and reflection layers.

\subsection{Dual-Stream Channel Reconfiguration Transformer}

DSCRT consists of three core components, namely GPFEN, TPFEN, and DPFIN. After receiving the transmission prior $\hat{T}_{\text{prior}}$ generated by LLCN, TPFEN processes both the mixed image $I$ and $\hat{T}_{\text{prior}}$ as dual-stream inputs, extracting initial features via $3 \times 3$ convolutions at the stem layer:
\begin{equation}
	F_t^{0,l} = \text{Conv}_{3\times3}(I), \quad F_t^{0,r} = \text{Conv}_{3\times3}(\hat{T}_{\text{prior}}),
\end{equation}
where $F_t^{0,l}$ denotes the convolutional prior features and $F_t^{0,r}$ denotes the transmission prior features. These dual-stream features then undergo multi-stage downsampling, with each stage incorporating MuGI blocks and downsampling convolutional layers, thereby forming multi-scale transmission prior features $(F_t^0, F_t^1, F_t^2, F_t^3, F_t^4, F_t^5)$, where each layer encompasses left and right dual streams $(F_t^{i,l}, F_t^{i,r})$, with $i$ representing the scale level index. Concurrently, GPFEN leverages a pre-trained Swin Transformer to extract multi-scale general prior features $(F_g^2, F_g^3, F_g^4, F_g^5)$ from $I$.

Subsequently, DPFIN conducts hierarchical feature fusion. At the 5-th layer, the general prior features and transmission prior features are processed through PixelShuffle operations and subsequently fused at the same layer via DSCRAB for same-layer fusion:
\begin{equation}
	F_{\text{same}}^{5,l} = \text{DSCRAB}(\text{PixelShuffle}(F_g^5), \text{PixelShuffle}(F_t^{5,l})),
\end{equation}
\begin{equation}
	F_{\text{same}}^{5,r} = \text{DSCRAB}(\text{PixelShuffle}(F_g^5), \text{PixelShuffle}(F_t^{5,r})).
\end{equation}

Similarly, the 4-th layer obtains same-layer fusion prior features $(F_{\text{same}}^{4,l}, F_{\text{same}}^{4,r})$ through an analogous process. Following this, the same-layer fusion prior features from the 5-th layer are refined through convolution and merged with those of the 4-th layer via DSCRAB to achieve cross-layer fusion:
\begin{equation}
	F_{\text{cross}}^{4,l} = \text{DSCRAB}(\text{Conv}(F_{\text{same}}^{5,l}, F_{\text{same}}^{5,r}), F_{\text{same}}^{4,l}),
\end{equation}
\begin{equation}
	F_{\text{cross}}^{4,r} = \text{DSCRAB}(\text{Conv}(F_{\text{same}}^{5,l}, F_{\text{same}}^{5,r}), F_{\text{same}}^{4,r}).
\end{equation}

The cross-layer fusion prior features from the 4-th layer are sequentially processed through PixelShuffle upsampling, MuGI block dual-stream interaction, and convolution operations. Subsequently, they are fused with the same-layer fusion prior features of the 3-rd layer through DSCRAB, generating the cross-layer fusion prior features of the 3-rd layer. This hierarchical propagation continues upward layer by layer until reaching the 2-nd layer. At the 1-st layer, given the absence of corresponding general prior features, the upper-layer cross-layer fusion prior features are directly merged with the current layer's transmission prior features. At the 0-th layer, the fusion encompasses the upper-layer cross-layer fusion prior features, current layer's transmission prior features, and convolutional prior features.

At the original resolution level, the network conducts MuGI dual-stream interaction on the 0-th layer's fusion prior features, subsequently outputting the transmission layer and reflection layer separately through convolutional layers:
\begin{equation}
	\hat{T}, \hat{R} = \text{Conv}(\text{MuGI}(F_{\text{cross}}^{0,l}, F_{\text{cross}}^{0,r})).
\end{equation}

In parallel, the LRM module estimates the nonlinear residual term:
\begin{equation}
	\hat{\Phi} = \text{LRM}(\text{MuGI}(F_{\text{cross}}^{0,l}, F_{\text{cross}}^{0,r})).
\end{equation}

\begin{table*}[t]
	\centering
	\caption{Quantitative Comparison Between Our Method and State-of-the-Art Methods on Five Real-World Testing Datasets. The Best Result Is Displayed in \textbf{Bold}, and the Second-Best Is \underline{Underlined}. $\star$ Denotes Using Official Pretrained Weights.}
	
	\label{tab:sirs_results}
	\scalebox{1}{
		\begin{tabular}{ccccccccccccc}
			\toprule
			\multirow{2}{*}{Methods} & \multicolumn{2}{c}{Real20 (20)} & 
			\multicolumn{2}{c}{Objects (200)} & \multicolumn{2}{c}{Postcard (199)} & \multicolumn{2}{c}{Wild (55)} & \multicolumn{2}{c}{Nature (20)} & \multicolumn{2}{c}{Average (494)} \\
			\cmidrule(lr){2-3} \cmidrule(lr){4-5} \cmidrule(lr){6-7} \cmidrule(lr){8-9} \cmidrule(lr){10-11} \cmidrule(lr){12-13}
			& PSNR$\uparrow$ & SSIM$\uparrow$ & PSNR$\uparrow$ & SSIM$\uparrow$ & PSNR$\uparrow$ & SSIM$\uparrow$ & PSNR$\uparrow$ & SSIM$\uparrow$ & PSNR$\uparrow$ & SSIM$\uparrow$ & PSNR$\uparrow$ & SSIM$\uparrow$ \\
			\midrule
			Li et al. \cite{li2020single}$\star$ & 21.85 & 0.777 & 24.95 & 0.899 & 23.45 & 0.879 & 24.35 & 0.886 & 24.03 & 0.798 & 24.11 & 0.881 \\
			Dong et al. \cite{dong2021location}$\star$ & 23.08 & 0.826 & 24.16 & 0.899 & 24.27 & 0.907 & 26.03 & 0.900 & 23.66 & 0.819 & 24.35 & 0.896 \\
			DSRNet \cite{hu2023single}$\star$ & 23.85 & 0.809 & 26.88 & 0.923 & 24.72 & 0.915 & 27.04 & 0.915 & 25.27 & 0.836 & 25.84 & 0.910 \\
			HGNet \cite{zhu2023hue}$\star$ & 23.78 & 0.818 & 25.11 & 0.902 & 23.85 & 0.900 & 27.05 & 0.900 & 25.51 & 0.827 & 24.78 & 0.895 \\
			Zhu et al. \cite{zhu2024revisiting}$\star$ & 21.93 & 0.788 & 26.89 & 0.925 & 24.29 & 0.887 & 26.82 & 0.910 & 26.14 & 0.846 & 25.60 & 0.899 \\
			DSIT \cite{hu2024single}$\star$ & 25.19 & 0.834 & 26.87 & 0.925 & \underline{26.38} & \underline{0.925} & \underline{27.90} & \underline{0.923} & \underline{26.68} & \underline{0.847} & 26.71 & \underline{0.918} \\
			RDNet \cite{zhao2025reversible}$\star$ & \textbf{25.71} & \textbf{0.850} & \underline{26.95} & \underline{0.926} & 26.33 & 0.922 & 27.84 & 0.917 & 26.31 & 0.846 & \underline{26.72} & 0.917 \\
			\midrule
			LLCN & 23.80 & 0.805 & 26.67 & 0.916 & 25.46 & 0.895 & 27.21 & 0.907 & 26.49 & 0.827 & 26.12 & 0.899 \\
			DPIT & \underline{25.46} & \underline{0.844} & \textbf{27.38} & \textbf{0.931} & \textbf{26.98} & \textbf{0.932} & \textbf{28.11} & \textbf{0.926} & \textbf{27.15} & \textbf{0.860} & \textbf{27.21} & \textbf{0.924} \\
			\bottomrule
		\end{tabular}
	}
\end{table*}

\begin{table*}[t]
	\centering
	\caption{Efficiency and Performance Comparison of Our Method with State-of-the-Art Methods on Five Real-World Testing Datasets. The Best Result Is Displayed in \textbf{Bold}, and the Second-Best Is \underline{Underlined}. $\star$ Denotes Using Official Pretrained Weights.}
	
	\label{tab:method_efficiency}
	\scalebox{1}{
		\begin{tabular}{ccccccc}
			\toprule
			\multirow{2}{*}{Methods} & \multirow{2}{*}{Venue} & \multicolumn{2}{c}{Efficiency} & \multicolumn{2}{c}{Performance} \\
			\cmidrule(lr){3-4} \cmidrule(lr){5-6}
			& & Trainable Params(M)$\downarrow$ & FLOPs(G)$\downarrow$ & Avg PSNR$\uparrow$ & Avg SSIM$\uparrow$ \\
			\midrule
			Li et al. \cite{li2020single}$\star$ & CVPR 2020 & 21.61 & 300.35 & 24.11 & 0.881 \\
			Dong et al. \cite{dong2021location}$\star$ & ICCV 2021 & \textbf{10.93} & 256.11 & 24.35 & 0.896 \\
			DSRNet \cite{hu2023single}$\star$ & ICCV 2023 & 123.67 & 276.59 & 25.84 & 0.910 \\
			HGNet \cite{zhu2023hue}$\star$ & TNNLS 2023 & \underline{14.50} & 303.75 & 24.78 & 0.895 \\
			Zhu et al. \cite{zhu2024revisiting}$\star$ & CVPR 2024 & 19.67 & \textbf{12.33} & 25.60 & 0.899 \\
			DSIT \cite{hu2024single}$\star$ & NeurIPS 2024 & 131.76 & 233.09 & 26.71 & \underline{0.918} \\
			RDNet \cite{zhao2025reversible}$\star$ & CVPR 2025 & 315.89 & 183.90 & \underline{26.72} & 0.917 \\
			\midrule
			LLCN & - & 99.44 & \underline{24.10} & 26.12 & 0.899 \\
			DPIT  & - & 131.54 & 191.35 & \textbf{27.21} & \textbf{0.924} \\
			\bottomrule
		\end{tabular}
	}
	\vspace{-3mm}
\end{table*}

\subsection{Loss Function}
To ensure the spatial consistency between the estimated layers and their ground truths, we employ a pixel reconstruction loss based on the mean squared error. Note that the ground truth for the reflection layer is derived as $R = |I - T|$. The pixel reconstruction loss is formulated as:
\begin{equation}
	\mathcal{L}_{\text{pix}} = \|\hat{T} - T\|_2^2 + \|\hat{R} - R\|_2^2,
\end{equation}
where $\hat{T}$ and $\hat{R}$ denote the estimated transmission and reflection layers, $T$ and $R$ are their corresponding ground truths, and $\|\cdot\|_2$ denotes the $\ell_2$ norm.

To preserve the structural integrity of the separated layers, we further incorporate a gradient reconstruction loss:
\begin{equation}
	\mathcal{L}_{\text{grad}} = \|\nabla\hat{T} - \nabla T\|_1 + \|\nabla\hat{R} - \nabla R\|_1,
\end{equation}
where $\nabla$ represents the gradient operator and $\|\cdot\|_1$ denotes the $\ell_1$ norm.

To enhance perceptual quality, we utilize features from a pre-trained VGG-19 network to construct the perceptual loss:
\begin{equation}
	\mathcal{L}_{\text{per}} = \sum_{i} \omega_i\|\phi_i(\hat{T}) - \phi_i(T)\|_1,
\end{equation}
where $\phi_i(\cdot)$ extracts features at layer $i$ of the VGG-19 model, with $i \in \{2, 7, 12, 21, 30\}$ representing the selected layer indices, and $\omega_i$ is the corresponding weighting coefficient.

To enforce consistency in the layer separation process, we introduce a reconstruction loss with a learnable residual term:
\begin{equation}
	\mathcal{L}_{\text{rec}} = \|I - (\hat{T} + \hat{R}) - \hat{\Phi}(\hat{T}, \hat{R})\|_1,
\end{equation}
where $\hat{\Phi}$ represents the learnable nonlinear residual component. This term effectively captures information beyond the additive model, thereby improving the completeness and clarity of the separated layers.

The overall loss function is defined as a weighted combination of all aforementioned terms:
\begin{equation}
	\mathcal{L}_{\text{total}} = \lambda_1\mathcal{L}_{\text{pix}} + \lambda_2\mathcal{L}_{\text{grad}} + \lambda_3\mathcal{L}_{\text{per}} + \lambda_4\mathcal{L}_{\text{rec}},
\end{equation}
where $\lambda_1 = 1$, $\lambda_2 = 1$, $\lambda_3 = 0.01$, and $\lambda_4 = 0.2$ are empirically determined balancing coefficients.

\section{Experimental Validation}\label{sec:experiments}

The LLCN and DSCRT are configured with distinct training setups. Specifically, images are resized to $224 \times 224$ dimensions using a batch size of 2 with 2 gradient accumulation steps for LLCN, whereas DSCRT processes images at $384 \times 384$ resolution with a batch size of 1. Both architectures utilize the Adam optimization algorithm with $\beta_1=0.9$, $\beta_2=0.999$, and a learning rate of $10^{-4}$.

Our training procedure comprises two sequential phases. During the initial phase, LLCN and DSCRT undergo separate training over 80 epochs, with optimal model weights determined according to L1 loss performance on the validation dataset. Subsequently, in the second phase, the selected LLCN and DSCRT components are integrated into the unified DPIT framework and refined through an additional 20 epochs. All experimental procedures are conducted on a single NVIDIA RTX 4090 GPU.

\subsection{Data Preparation}

\subsubsection{Training Data}

Our training corpus incorporates both synthetically generated and authentic image pairs. Synthetic pairs are created from the PASCAL VOC database \cite{everingham2010pascal} employing the DSIT blending methodology, allocating 500 pairs for DPIT and 2000 pairs for LLCN training. The real image collection encompasses 89 pairs from Zhang et al.'s work \cite{zhang2018single} and 200 pairs sourced from the Nature dataset \cite{li2020single}.

We synthesize images $I_{\text{syn}}$ by combining transmission layer $T_{\text{syn}}$ and reflection layer $R_{\text{syn}}$ according to:
\begin{equation}
	I_{\text{syn}} = \gamma_1 T_{\text{syn}} + \gamma_2 R_{\text{syn}} - \gamma_1\gamma_2 T_{\text{syn}}  R_{\text{syn}}.
\end{equation}

Here, blending coefficients $\gamma_1 \in [0.8, 1.0]$ and $\gamma_2 \in [0.4, 1.0]$ modulate the contribution strengths of transmission and reflection components. We implement a hybrid sampling approach during training. Technically, it draws from Synthetic, Real, and Natural image pairs with a 0.6:0.2:0.2 distribution ratio, while maintaining a consistent sample size of 4,000 image pairs per epoch for both models.

\subsubsection{Validation Data}

We designate 50 image pairs from the Reflection Removal in the Wild (RRW) dataset \cite{zhu2024revisiting} to serve as our validation benchmark for tracking performance metrics and guiding model selection throughout the training process. RRW constitutes a comprehensive real-world reflection removal dataset featuring authentic scenes photographed through glass surfaces, encompassing varied indoor and outdoor settings with diverse illumination characteristics, thereby facilitating robust assessment of model generalization capabilities.

\subsubsection{Test Data}

We assess model efficacy across five real-world benchmark datasets: Real, Nature, and three SIR$^2$ subsets \cite{wan2017benchmarking}. The Real evaluation set comprises 20 image pairs captured through portable glass panels across various indoor and outdoor contexts. The Nature evaluation set consists of 20 authentic reflection image pairs from natural environments. The SIR$^2$ collection is divided into three categories: the Objects subset features 200 indoor household item pairs, such as ceramic mugs, plush toys, and fruits. The Postcard subset presents 199 controlled scenario pairs derived from pairwise postcard combinations. The Wild subset encompasses 55 outdoor scene pairs showcasing complex compositions with tree foliage, glass windows, and architectural structures. These evaluation datasets provide substantial diversity in scene composition, lighting configurations, and reflection properties, enabling thorough validation of reflection removal effectiveness.

\subsection{Performance Evaluation}

This section establishes the superior performance of DPIT in addressing the single image reflection removal challenge. Our comparative analysis encompasses seven cutting-edge approaches: Li et al. \cite{li2020single}, Dong et al. \cite{dong2021location}, DSRNet \cite{hu2023single}, HGNet \cite{zhu2023hue}, Zhu et al. \cite{zhu2024revisiting}, DSIT \cite{hu2024single}, and RDNet \cite{zhao2025reversible}. Additionally, we present LLCN results to showcase the effectiveness of lightweight transmission prior generation. Performance assessment is carried out across five real-world datasets, namely Real test set \cite{zhang2018single}, Objects \cite{wan2017benchmarking}, Postcard \cite{wan2017benchmarking}, Wild \cite{wan2017benchmarking}, and Nature test set \cite{li2020single}, employing Peak Signal-to-Noise Ratio (PSNR) and Structural Similarity Index (SSIM) \cite{wang2004image} as quantitative metrics. The computational complexity in terms of FLOPs is evaluated with inputs of size $224 \times 224$ pixels.

\textit{1) Quantitative Comparison:} Table~\ref{tab:sirs_results} summarizes the quantitative performance of various methods across five benchmark datasets. DPIT demonstrates state-of-the-art results, achieving an average PSNR of 27.21 dB and SSIM of 0.924. The method achieves top performance on four out of five datasets: 27.38 dB/0.931 for Objects, 26.98 dB/0.932 for Postcard, 28.11 dB/0.926 for Wild, and 27.15 dB/0.860 for the Nature test set, exceeding the runner-up by margins of 0.43 dB/0.005, 0.60 dB/0.007, 0.21 dB/0.003, and 0.47 dB/0.013, respectively. For the Real test set, DPIT achieves 25.46 dB/0.844, marginally trailing RDNet's 25.71 dB/0.850, yet maintaining a substantial advantage in overall performance. Such consistently outstanding results across varied scenarios featuring diverse reflection properties and lighting environments confirm the method's robustness and strong generalization capacity.

Table~\ref{tab:method_efficiency} delivers an integrated comparison considering both efficiency and performance aspects. DPIT strikes an optimal trade-off through the streamlined architecture of the Local Linear Correction Network and the computational efficiency of the Dual-Stream Channel Reorganization Attention mechanism. Concretely, DPIT delivers 27.21 dB/0.924 utilizing 131.54M trainable parameters and 191.35G FLOPs. Relative to RDNet's 315.89M trainable parameters, DPIT operates with merely 41.6\% of the parameter count while yielding a performance gain of 0.49 dB/0.007. When compared against DSIT, which similarly leverages attention mechanisms, DPIT demonstrates reduced computational overhead even with the added transmission prior component, requiring 191.35G FLOPs versus DSIT's 233.09G, reflecting a 17.9\% reduction, while concurrently enhancing performance by 0.50 dB/0.006. Against DSRNet, DPIT secures a performance boost of 1.37 dB/0.014 while decreasing FLOPs from 276.59G to 191.35G, corresponding to a 30.8\% reduction. These findings comprehensively illustrate the substantial efficiency gains achieved by our approach alongside performance improvements.

Component-wise analysis reinforces the design's efficacy. LLCN achieves 26.12 dB/0.899 with only 99.44M trainable parameters and 24.10G FLOPs, exhibiting exceptional parameter efficiency and presenting a compelling blueprint for high-quality lightweight reflection removal networks. The integrated DPIT architecture achieves profound fusion of the transmission and general priors via the dual-prior interaction mechanism, yielding a 1.09 dB/0.025 improvement over LLCN. This substantiates the considerable merit of harnessing complementary information from both priors for reflection removal.

\begin{figure*}[t]
	\centering
	\includegraphics[width=1\linewidth]{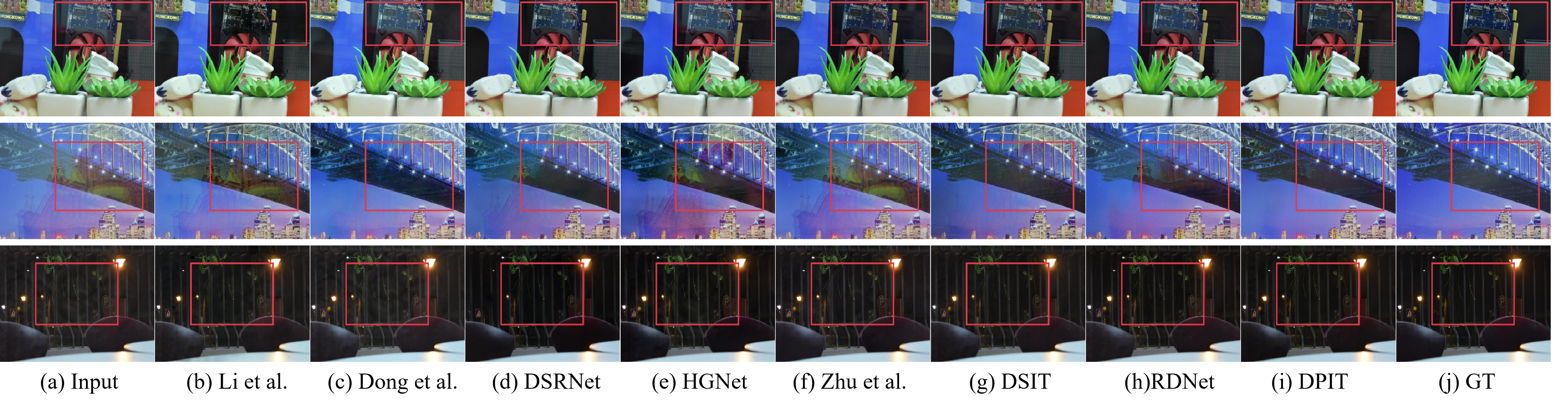}
\caption{Comparison of single image reflection removal results by different methods on samples from Objects, Postcard, and Wild datasets. From top to bottom are Objects, Postcard, and Wild.}	
	\label{fig:objects_postcard_wild}	
\end{figure*}

\begin{figure*}[t]
	\centering
	\includegraphics[width=1\linewidth]{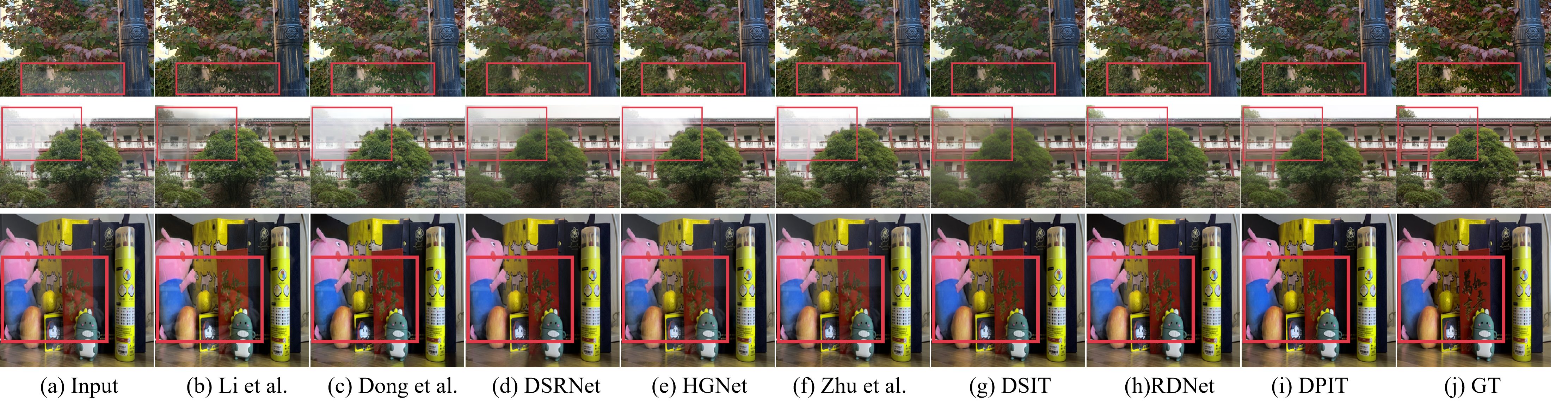}
\caption{Comparison of single image reflection removal results by different methods on the Real test set, Nature test set, and Reflection Removal in the Wild (RRW). From top to bottom are Real, Nature, and RRW.}
	\label{fig:real20_nature_rrw}	
\end{figure*}

\begin{figure*}[t]
	\centering
	\includegraphics[width=1\linewidth]{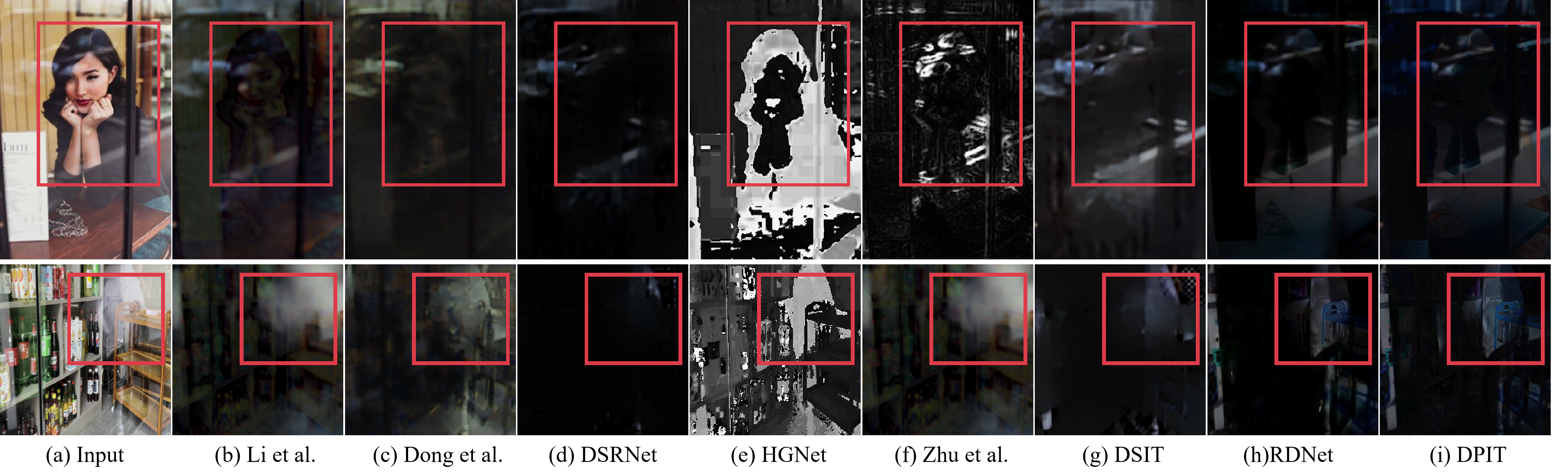}
\caption{Comparison of reflection or other non-transmission component separation results by different methods on the Real45 and Reflection Removal in the Wild (RRW) datasets. From top to bottom are Real45 and RRW.}
	\label{fig:real45}	
\end{figure*}

\textit{2) Qualitative Comparison:} To provide a more intuitive demonstration of performance differences across methods, this section evaluates reflection removal effectiveness through visual comparisons on multiple real-world scene datasets.

Figure~\ref{fig:objects_postcard_wild} illustrates transmission layer recovery outcomes achieved by different methods on the Objects, Postcard, and Wild datasets. In the bridge scene from the Postcard dataset, reflections are distributed primarily across three regions: bridge railings, bridge back surface, and sky background. The comparative results reveal distinct performance patterns. The method by Li et al. \cite{li2020single} retains substantial reflection components in the bridge back and sky regions. Dong et al.'s approach \cite{dong2021location} exhibits noticeable residuals in the bridge railing area. DSRNet \cite{hu2023single} displays visible reflection traces in both bridge railings and sky background. HGNet \cite{zhu2023hue} fails to completely eliminate reflections in the bridge railing and sky regions. The work of Zhu et al. \cite{zhu2024revisiting} demonstrates varying degrees of residuals across all three main regions. DSIT \cite{hu2024single} continues to show visible reflections in the bridge railing area, while RDNet \cite{zhao2025reversible} exhibits residuals on the bridge back surface. In contrast, DPIT achieves nearly complete removal across these three reflection-concentrated regions, producing visual results closest to the ground truth transmission layer. The indoor object scene from the Objects dataset further validates DPIT's superiority, achieving thorough reflection suppression while maintaining texture clarity and achieving an ideal balance between reflection removal and detail preservation. The nighttime restaurant scene from the Wild dataset presents non-uniformly distributed reflections under low-light conditions. DPIT effectively removes these reflection components while accurately preserving scene structures and textural details.

Figure~\ref{fig:real20_nature_rrw} presents transmission layer recovery outcomes on the Real test set, Nature test set, and RRW validation set. In the building scene from the Nature test set, reflections concentrate primarily in the attic region at the upper left corner. The comparative results show that the methods by Dong et al. \cite{dong2021location}, HGNet \cite{zhu2023hue}, and Zhu et al. \cite{zhu2024revisiting} leave extensive reflection residuals in that region. DSRNet \cite{hu2023single} shows improvement, yet building details within the reflection region remain blurred. The approaches by Li et al. \cite{li2020single} and DSIT \cite{hu2024single} suffer from texture detail loss due to over-smoothing, while RDNet \cite{zhao2025reversible}'s output exhibits artifacts. DPIT successfully removes reflections in the attic region while effectively maintaining the clarity and detail integrity of the building structure, demonstrating exceptional capability in complex building scenes. The Real test set contains scenes with leaves and pillars. Although the relatively weak reflection intensity enables most methods to achieve acceptable results, DPIT still performs better in preserving fine structures. The RRW validation set presents indoor multi-object scenes where the complex mixture of transmission and reflection layers increases separation difficulty, yet DPIT effectively removes reflection components and accurately restores scene details, demonstrating excellent generalization performance.

Figure~\ref{fig:real45} compares reflection layer separation outcomes on the Real45 and RRW validation sets. Since these datasets lack ground truth annotations, this comparison better reflects the actual capabilities and generalization performance of various methods in reflection component extraction. In the portrait scene from the Real45 dataset, the reflection layer extracted by DPIT appears clear and complete, presenting optimal separation quality. The convenience store scene from the RRW validation set presents greater challenges. The methods by Li et al. \cite{li2020single} and Zhu et al. \cite{zhu2024revisiting} can only extract blurred highlight regions lacking clear structural information. DSRNet \cite{hu2023single}'s output appears nearly black, indicating insufficient extraction capability. The approaches by Dong et al. \cite{dong2021location} and DSIT \cite{hu2024single} can capture general reflection contours but exhibit severe loss of internal details. Both RDNet \cite{zhao2025reversible} and DPIT achieve relatively successful separation, clearly presenting the structural characteristics of the reflection layer. Among these, DPIT demonstrates particularly outstanding performance in terms of detail richness, brightness distribution, and structural integrity. It not only achieves strong results in transmission layer recovery tasks but also demonstrates excellent performance in reflection layer separation tasks, fully validating the universality and effectiveness of the proposed method across different tasks.

\begin{table*}[t]
	\centering
\caption{Ablation Study Comparing Different Modeling Methods for the LLCN, Showing Both Efficiency and Performance Metrics. The Best Result Is Displayed in \textbf{Bold}, and the Second-Best Is \underline{Underlined}.}
	\label{tab:ablation}
	\scalebox{0.8}{
		\begin{tabular}{ccccccccccccccccc}
			\toprule
			\multirow{3}{*}{Modeling Method} & \multicolumn{2}{c}{Efficiency} & \multicolumn{2}{c}{Real20 (20)} & \multicolumn{2}{c}{Objects (200)} & \multicolumn{2}{c}{Postcard (199)} & \multicolumn{2}{c}{Wild (55)} & \multicolumn{2}{c}{Nature (20)} & \multicolumn{2}{c}{Average (494)} \\
			\cmidrule(lr){2-3} \cmidrule(lr){4-5} \cmidrule(lr){6-7} \cmidrule(lr){8-9} \cmidrule(lr){10-11} \cmidrule(lr){12-13} \cmidrule(lr){14-15}
			& Params(M)$\downarrow$ & FLOPs(G)$\downarrow$ & PSNR$\uparrow$ & SSIM$\uparrow$ & PSNR$\uparrow$ & SSIM$\uparrow$ & PSNR$\uparrow$ & SSIM$\uparrow$ & PSNR$\uparrow$ & SSIM$\uparrow$ & PSNR$\uparrow$ & SSIM$\uparrow$ & PSNR$\uparrow$ & SSIM$\uparrow$\\
			\midrule
			$T = f_\theta(I)$ & \underline{93.50} & \underline{19.73} & 22.45 & 0.721 & 25.30 & 0.857 & 23.02 & 0.799 & 26.03 & 0.866 & \underline{26.07} & 0.815 & 24.38 & 0.827 \\
			$I = T + R + \Phi(T, R)$ & 105.38 & 28.47 & 22.91 & 0.704 & 25.33 & 0.845 & 23.24 & 0.787 & 25.86 & 0.857 & 25.10 & 0.795 & 24.44 & 0.816 \\
			$I = sI + b + R + \Phi(T, R)$ & 111.32 & 32.84 & \underline{23.67} & \underline{0.797} & \underline{26.27} & \underline{0.911} & \underline{24.61} & \underline{0.888} & \underline{26.86} & \underline{0.902} & 25.92 & \underline{0.820} & \underline{25.55} & \underline{0.892} \\
			$T = \alpha I + \beta$ & \textbf{87.57} & \textbf{15.35} & 21.10 & 0.755 & 25.58 & 0.897 & 23.45 & 0.885 & 26.38 & 0.896 & 22.16 & 0.784 & 24.49 & 0.882 \\
			$T = sI + b$ & 99.44 & 24.10 & \textbf{23.80} & \textbf{0.805} & \textbf{26.67} & \textbf{0.916} & \textbf{25.46} & \textbf{0.895} & \textbf{27.21} & \textbf{0.907} & \textbf{26.49} & \textbf{0.827} & \textbf{26.12} & \textbf{0.899} \\
			\bottomrule
		\end{tabular}
	} 
\end{table*}

\begin{table*}[t]
	\centering
\caption{Ablation Study on DPIT with Different Dual-Stream Interaction Blocks with and without Transmission Prior, Showing Both Efficiency and Performance Metrics. The Best Results Is Displayed in \textbf{Bold}, and the Second-Best Is \underline{Underlined}.}
	\label{tab:ablation2}
	\scalebox{0.81}{
		\begin{tabular}{cccccccccccccccc}
			\toprule
			\multirow{3}{*}{\shortstack{Dual-Stream\\Interaction Block}} & \multirow{3}{*}{\shortstack{Trans\\Prior}} & \multicolumn{2}{c}{Efficiency} & \multicolumn{12}{c}{Performance} \\
			\cmidrule(lr){3-4} \cmidrule(lr){5-16}
			& & \multirow{2}{*}{Params(M)$\downarrow$} & \multirow{2}{*}{FLOPs(G)$\downarrow$} & \multicolumn{2}{c}{Real20 (20)} & \multicolumn{2}{c}{Objects (200)} & \multicolumn{2}{c}{Postcard (199)} & \multicolumn{2}{c}{Wild (55)} & \multicolumn{2}{c}{Nature (20)} & \multicolumn{2}{c}{Average (494)} \\
			\cmidrule(lr){5-6} \cmidrule(lr){7-8} \cmidrule(lr){9-10} \cmidrule(lr){11-12} \cmidrule(lr){13-14} \cmidrule(lr){15-16}
			& & & & PSNR$\uparrow$ & SSIM$\uparrow$ & PSNR$\uparrow$ & SSIM$\uparrow$ & PSNR$\uparrow$ & SSIM$\uparrow$ & PSNR$\uparrow$ & SSIM$\uparrow$ & PSNR$\uparrow$ & SSIM$\uparrow$ & PSNR$\uparrow$ & SSIM$\uparrow$\\
			\midrule
			MLP & w/o & 168.30 & \underline{140.90} & 23.89 & 0.827 & 25.98 & 0.915 & 23.61 & 0.886 & 26.58 & 0.911 & 25.86 & 0.846 & 25.00 & 0.896 \\
			YTMT & w/o & 444.16 & 254.65 & 24.05 & 0.826 & 26.32 & 0.920 & 24.17 & 0.902 & 27.11 & 0.913 & 25.87 & 0.847 & 25.43 & 0.905 \\
			MuGI & w/o & \textbf{84.51} & \textbf{125.58} & 25.38 & 0.835 & 26.88 & 0.928 & 25.09 & 0.918 & 27.18 & 0.916 & 27.22 & 0.852 & 26.15 & 0.916 \\
			DAIB & w/o & 131.76 & 233.09 & \underline{25.46} & \underline{0.841} & 27.01 & 0.928 & 25.77 & \underline{0.930} & 27.32 & 0.917 & 27.12 & \textbf{0.860} & 26.49 & \underline{0.922} \\
			DSCRAB & w/o & \underline{131.54} & 167.25 & 24.96 & 0.834 & 27.03 & 0.926 & \underline{26.64} & 0.928 & 27.57 & 0.920 & 26.96 & 0.855 & 26.85 & 0.919 \\
			\cmidrule(lr){1-16}
			MLP & w/ & 168.30 & 164.99 & 24.73 & 0.834 & 26.72 & 0.924 & 25.91 & 0.915 & 27.58 & 0.919 & 26.70 & \underline{0.857} & 26.41 & 0.913 \\
			YTMT & w/ & 444.16 & 278.75 & 25.13 & 0.839 & \textbf{27.48} & \textbf{0.932} & 26.02 & 0.915 & \textbf{28.16} & 0.922 & 27.02 & \underline{0.857} & 26.85 & 0.917 \\
			MuGI & w/ & \textbf{84.51} & 149.67 & \textbf{25.49} & 0.840 & 27.17 & 0.930 & 26.20 & 0.921 & 27.42 & 0.918 & \underline{27.33} & \underline{0.857} & 26.75 & 0.918 \\
			DAIB & w/ & 131.76 & 257.19 & 25.15 & \underline{0.841} & 27.27 & \underline{0.931} & 26.39 & 0.926 & 27.89 & \underline{0.923} & \textbf{27.40} & \textbf{0.860} & \underline{26.90} & 0.921 \\
			DSCRAB & w/ & \underline{131.54} & 191.35 & \underline{25.46} & \textbf{0.844} & \underline{27.38} & \underline{0.931} & \textbf{26.98} & \textbf{0.932} & \underline{28.11} & \textbf{0.926} & 27.15 & \textbf{0.860} & \textbf{27.21} & \textbf{0.924} \\
			\bottomrule
		\end{tabular}
	} 
\end{table*}

\subsection{Ablation Study}

\textit{1) Ablation Study on Modeling Methods for Lightweight Transmission Prior Generation Network:} To verify the effectiveness of the proposed local linear correction modeling approach, we conducted systematic ablation experiments comparing five different modeling strategies. Table~\ref{tab:ablation} presents detailed quantitative results, including computational efficiency and performance on five test datasets. To ensure fair comparison, all methods employ the pre-trained ConvNeXt-Base as the feature extraction backbone with a unified decoder structural design. The only difference is the number of decoders, which varies based on the requirements of each modeling method.

We first examine methods that directly generate the transmission layer. The method ${T} = f_\theta({I})$ uses a single decoder to decode the last-layer features of ConvNeXt-Base, directly generating the transmission layer image, achieving 24.38 dB/0.827 with 93.50M parameters and 19.73G FLOPs. The method ${I} = {T} + {R} + \Phi({T}, {R})$ employs three parallel decoders, similarly decoding the last-layer features to generate the transmission layer, reflection layer, and their nonlinear coupling term, requiring 105.38M parameters and 28.47G FLOPs while achieving 24.44 dB/0.816. These results demonstrate that directly generating complete images under parameter constraints fails to achieve ideal results. Even with multiple decoders modeling the complete degradation process, performance improvement remains limited.

Next, we examine methods based on linear correction. The method ${T} = \alpha{I} + \beta$ uses two decoders to predict global scaling coefficient $\alpha$ and bias coefficient $\beta$, achieving 24.49 dB/0.882 with 87.57M parameters and 15.35G FLOPs, demonstrating optimal computational efficiency. This method estimates the transmission layer by performing linear correction on the input image, which shares a similar concept with our approach, yet the globally uniform transformation lacks fine-grained spatial and channel-wise adaptability. In contrast, our proposed local linear correction model ${T} = s{I} + b$ also uses two decoders to predict pixel-wise and channel-wise scaling factor $s$ and bias term $b$, achieving the best performance of 26.12 dB/0.899 with 99.44M parameters and 24.10G FLOPs. Specifically, it achieves 23.80 dB/0.805 on Real, 26.67 dB/0.916 on Objects, 25.46 dB/0.895 on Postcard, 27.21 dB/0.907 on Wild, and 26.49 dB/0.827 on Nature. Local linear correction improves PSNR and SSIM by 1.63 dB and 0.017 over global linear transformation, thereby validating the necessity of spatially adaptive modeling.

To further verify the superiority of the local linear correction modeling approach, the method ${I} = s{I} + b + {R} + \Phi({T}, {R})$ replaces the transmission layer generation in complete degradation modeling with our modeling formula, using four decoders to predict $s$, $b$, ${R}$, and $\Phi$. This method requires 111.32M parameters and 32.84G FLOPs, with performance significantly improving to 25.55 dB/0.892. Compared to the method ${I} = {T} + {R} + \Phi({T}, {R})$ with 24.44 dB/0.816, merely replacing the transmission layer modeling approach yields significant improvements of 1.11 dB and 0.076 in PSNR and SSIM, fully demonstrating the effectiveness of local linear correction modeling.

Comprehensive analysis shows that the proposed local linear correction model achieves advantages in multiple aspects. Compared to the end-to-end baseline, the proposed model improves PSNR and SSIM by 1.74 dB and 0.072, from 24.38 dB/0.827 to 26.12 dB/0.899. Compared to global linear transformation, it improves PSNR and SSIM by 1.63 dB and 0.017, from 24.49 dB/0.882 to 26.12 dB/0.899. Compared to complete degradation modeling, it improves PSNR and SSIM by 0.57 dB and 0.007, from 25.55 dB/0.892 to 26.12 dB/0.899, while reducing parameters from 111.32M to 99.44M and computational cost from 32.84G FLOPs to 24.10G FLOPs. These results validate the core design principle of LLCN, which is to constrain the network to learn adaptive local corrections rather than directly generating the complete transmission layer, thereby achieving high-quality transmission prior generation within a compact parameter budget.

\textit{2) Performance Impact Analysis of Dual-Stream Interaction Modules:} To systematically evaluate the effectiveness of transmission prior and verify the performance advantages of the proposed dual-stream channel reorganization attention mechanism, we designed comparative experiments by testing DPIT with five different dual-stream interaction modules, including MLP, YTMT \cite{hu2021trash}, MuGI, DAIB \cite{hu2024single}, and our proposed DSCRAB, which serves as the default configuration. Each module is evaluated both with and without transmission prior integration, resulting in ten configurations to quantify the independent and combined contributions of each component. Table~\ref{tab:ablation2} presents detailed experimental results. The training adopts a two-stage strategy. In the first stage, LLCN is trained independently for 80 epochs, and each DPIT variant with different interaction modules is also trained separately for 80 epochs without using transmission prior. Optimal weights are selected based on L1 loss on the validation set. In the second stage, the selected LLCN is integrated with each corresponding DPIT variant to form complete models, which continue training for 20 epochs to enable transmission prior utilization.

In configurations without transmission prior, DPIT with MuGI as the dual-stream interaction module requires 84.51M parameters and 125.58G FLOPs, achieving 26.15 dB/0.916. Our proposed DSCRAB significantly improves performance to 26.85 dB/0.919, representing a 0.70 dB PSNR gain over MuGI, with 131.54M parameters and 167.25G FLOPs. Notably, DAIB \cite{hu2024single} achieves 26.49 dB/0.922 with 131.76M parameters and 233.09G FLOPs. Compared to DAIB, which has a similar parameter count, DSCRAB not only achieves 0.36 dB higher PSNR but also reduces computational cost by 28.2\%. These results demonstrate that the proposed dual-stream channel reorganization attention mechanism achieves superior performance and maintains competitive computational efficiency.

After introducing transmission prior, all configurations show performance improvements, demonstrating the effectiveness of transmission prior. Specifically, MuGI improves by 0.60 dB to 26.75 dB/0.918, DAIB improves by 0.41 dB to 26.90 dB/0.921, and DSCRAB improves by 0.36 dB to 27.21 dB/0.924, achieving the best performance in both PSNR and SSIM. Notably, introducing transmission prior brings performance improvements ranging from 0.36 dB to 1.42 dB, while the additional computational overhead is approximately 24.10G FLOPs, indicating that significant performance improvements can be obtained at a relatively small additional cost. Performance analysis across test sets shows that DPIT with transmission prior achieves the best results on all five test sets, with 25.46 dB/0.844 on Real, 27.38 dB/0.931 on Objects, 26.98 dB/0.932 on Postcard, 28.11 dB/0.926 on Wild, and 27.15 dB/0.860 on Nature. This consistent performance across datasets demonstrates that the proposed method exhibits strong generalization capability and robustness in reflection scenarios with different complexities and degradation patterns.

\section{Conclusion}\label{sec:conclusion}

We propose DPIT to effectively integrate transmission and general priors for single image reflection removal. We introduce LLCN based on the LLCM formulated as $T = sI + b$ for lightweight transmission prior generation, and design DSCRAB with DSCRAM for efficient dual-prior interaction and layer separation.

Extensive experiments validate the effectiveness of our approach. DPIT achieves 27.21 dB PSNR and 0.924 SSIM on average across five benchmark datasets, surpassing existing methods. The main results demonstrate the following. 1) LLCN generates superior transmission prior with limited parameters, achieving 1.63 dB improvement over global linear methods. 2) DSCRAB achieves superior performance while reducing computational costs by 17.9\% compared to DSIT by optimizing the reconstruction target. 3) Across different dual-stream interaction modules, introducing transmission prior consistently contributes gains of 0.36 to 1.42 dB, which validates the importance of transmission prior modeling for reflection removal tasks.

Our future work will focus on extending the local linear correction paradigm to broader image restoration tasks and developing generalized multi-prior interaction mechanisms for handling diverse heterogeneous priors in low-level vision.

\bibliographystyle{IEEEtran}
\bibliography{TMM2025.bib}
\end{document}